\documentclass[runningheads]{llncs}
\usepackage[utf8]{inputenc}
\usepackage{hyperref}
\usepackage{color}
\usepackage{xcolor}
\usepackage{tabularx}
\usepackage{authblk}
\usepackage{hyperref}
\usepackage{graphicx}
\usepackage{subfig}
\usepackage{float}
\usepackage{adjustbox}
\usepackage{neuralnetwork}
\usepackage{tikz}
\usepackage{graphics, fancyhdr}
\usepackage{algorithmicx}
\usepackage[ruled,vlined]{algorithm2e}
\usepackage{adjustbox}
\usepackage{graphicx}
\begin{document}
\raggedbottom
\title{Incremental Feature Learning For Infinite Data}
\author{Armin Sadreddin\inst{1} \and
Samira Sadaoui\inst{1}}
\authorrunning{S. Sadreddin and S. Sadaoui}
\institute{Computer Science Department, University of Regina, Regina, Canada\\
\email{ArminSadreddin@uregina.ca}\\
\email{Samira.Sadaoui@uregina.ca}}

\maketitle              

\begin{abstract}
This study addresses the actual behavior of the credit-card fraud detection environment where financial transactions containing sensitive data must not be amassed in an enormous amount to conduct learning.  We introduce a new adaptive learning approach that adjusts frequently and efficiently to new transaction chunks; each chunk is discarded after each incremental training step. Our approach combines transfer learning and incremental feature learning.  The former improves the feature relevancy for subsequent chunks, and the latter, a new paradigm, increases accuracy during training by determining the optimal network architecture dynamically for each new chunk.  The architectures of past incremental approaches are fixed; thus, the accuracy may not improve with new chunks. We show the effectiveness and superiority of our approach experimentally on an actual fraud dataset.

\end{abstract}

\section{Introduction}
A tremendous volume of credit card transactions is conducted daily, especially with the COVID-19 pandemic. Nevertheless, this financial activity necessities many robust resources, in terms of CPU, RAM, storage and human expertise, to detect fraudulent payments.  For the year 2019, the Nelson Report estimated the worldwide financial loss from credit card fraud to \$28.65 billion ~\cite{CCFstat2019}. Over the last ten years, users that revealed at least one credit card fraud soared by 71\% in Canada ~\cite{CCFstat2021}.  Credit-card fraud classification is complex to tackle due to the following challenges:  {\bf (1)} transactions are generated continuously and speedily. In this data stream context, fraud must be detected in real-time to avoid losses on the customers' side, {\bf (2)} Machine Learning Algorithms (MLAs) require storing a very large number of transactions to conduct batch learning. However, financial transactions must not be accumulated in an enormous quantity by the fraud detection models because of data sensitivity and confidentiality, {\bf (3)} MLAs cannot adapt previous knowledge to newly available transactions to improve their accuracy, making the detection models obsolete and unreliable in the long run, and {\bf (4)} the architectures of all past incremental learning approaches are pre-determined \cite{DBLP:journals/ci/AnowarS21}, \cite{DBLP:conf/flairs/AnowarS20}, thus, the accuracy may not improve with new data. 

Most of the credit-card fraud detection studies employed conventional MLAs that are however inadequate for this specific environment. In the industry, a new fraud prediction system is created from scratch for every number of days to learn the new behavior ~\cite{DBLP:conf/dsaa/LebichotPBSHO20}. However, re-training is very time-consuming, and the learned knowledge is completely lost. To overcome the three challenges above, we develop a new adaptive learning algorithm that learns frequently and efficiently from transaction chunks or mini-batches.  According to the incremental training frequency, we decide how much data to collect in each chunk. More the frequency is shorter,  less sensitive data is accumulated. In our study, a chunk contains the payment transactions of one day, which is still large. We only process one chunk at a time in the short-term memory and discard it after each model adaptation, without storing it.

\medskip

The proposed adaptive approach combines transfer learning and incremental feature learning (a new paradigm). Thanks to transfer learning, we extract valuable features from the original ones and reuse the new features for the subsequent transaction chunks. For instance, in the image processing area, the first layers of the neural network extract fundamental features that can be reused in another image processing task. Following the same reasoning, we use the first layers to collect more beneficial features and then add a new network to utilize those features. By doing so, we take advantage of the previous chunks' knowledge. 

Our new Incremental Feature Learning (IFL) algorithm adapts gradually to the new transaction chunks by  {\bf (1)} preserving the previously learned knowledge and {\bf (2)} dynamically adjusting the network architecture for each new chunk to achieve the highest performance during training.  IFL expands the network topology by adding new hidden layers and units during each adaptation phase.  Determining the most suitable model's architecture leading to the best performance is a complex problem. Looking for the optimal number of units in a hidden layer is already tricky for static MLAs.  Our IFL approach adds hidden units one by one until the model does not converge anymore. Nevertheless, as we are changing the network architecture to increase the performance, we may over-fit the resulting model.  Hence, we utilize a validation chunk during training to avoid over-fitting after each extension.  More precisely, only the weights of the new hidden units are updated each time as the previous units are frozen to store the previous knowledge. Thus, the less computational time is required to conduct learning for each new chunk. In this way, our IFL approach will always outperform other incremental learning approaches as the former continuously adapts its architecture to reach optimal accuracy. The architecture is permanently fixed for past incremental approaches, and the accuracy may not improve when new training chunks are provided.  Developing such an algorithm is very challenging, requiring a deep investigation of the building blocks and libraries of MLA toolkits, such as creating a new hidden unit, adding a new connection, and freezing the weights of a old connection (not to be re-optimized).

\medskip

There is only one study that developed an IFL approach, but very elementary ~\cite{DBLP:journals/npl/GuanL01}. Still, this paper does not provide details on the algorithm design and its implementation. Instead, our paper presents the steps of our new IFL approach that learns progressively from newly available chunks. Through a concrete example, we illustrate step by step the sophisticated behavior of our approach. Moreover, using a real credit-card fraud dataset, we create training, testing, and validation chunks and handle the highly imbalanced learning problem. We build four fraud classification models for the experiments: (1) the initial model trained on the first-day transaction chunk, (2) the initial model trained on the second-day transaction chunk, (3) the re-fitted model trained on the second-day chunk, and (4) the final optimal model trained on the second day and produced with our new approach. We thoroughly evaluate and compare all these learned models on unseen data.  
\section{Related Work}
We review recent research on detecting credit card fraud and highlight its weaknesses. The majority of studies conducted batch learning, such as ~\cite{Najadat2020} that explored deep learning, like BiLSTM and BiGRU, and classical learning, such as Decision Tree, Ada Boosting, Logistic Regression, Random Forest, Voting and Naive Base. Since the fraud dataset is highly imbalanced, the authors adopted random under-sampling, over-sampling and SMOTE. The hybrid of over-sampling, BiLSTM and BiGRU lead to the highest accuracy.  Another work  ~\cite{nguyen2020deep} also assessed several MLAs, including LSTM, 2-D CNN, 1-D CNN, Random Forest, ANN and SVM, using different data sampling methods on three credit-card datasets. LSTM and 1-D CNN combined with SMOTE returned the best results.  We believe LSTM can be a good option for incremental learning since this algorithm can remember past data and therefore creates predictions using the current inputs and past data, leading to a better response to the environmental changes.  In both papers, LSTMs and the other models were trained on very large datasets, requiring storing sensitive information forever.  Nevertheless, since user transactions are available incrementally, conventional MLAs are inappropriate for streaming data. Our proposed method aims to address the real credit-card fraud classification context. 

\smallskip 

In \cite{DBLP:conf/smc/AnowarS20}, the authors first utilized SMOTE-ENN to handle a highly imbalanced credit-card fraud dataset and then divided the dataset into multiple training chunks to simulate incoming data. They proposed an ANN-based incremental learning approach that learns gradually from new chunks using an incremental memory model. For adjusting the model each time, the memory consists of one past chunk (so that data are not forgotten immediately) and one recent chunk (to conduct the model adaptation). The authors demonstrated that incremental learning is superior to static learning. However, using two chunks every time can be expensive computationally. Also, since the ANN topology is fixed, the model cannot adapt to significant changes in the chunk patterns. In our study, the ANN architecture is dynamic to build an optimal fraud detection model. Instead of using two chunks simultaneously, leading to storing more data, we use only one chunk. With transfer learning, we take advantage of the previous chunk without storing it.

\smallskip 

In ~\cite{Barris2020}, the authors introduced an incremental Gradient Boosting Trees (GBT) approach, which is just an ensemble of decision trees, to minimize the loss function gradually. The ensemble is updated for each new group of transactions by appending a new Decision Tree to create a more robust GBT model.  The authors developed three models:  static GBT (batch learning), re-trained GBT (re-training all the transactions by including the
investigated ones), and incremental GBT (ensemble).  All these methods necessitate storing a tremendous quantity of user transactions. The authors divided the credit-card dataset into several sub-datasets w.r.t time (month) and evaluated the three models for each month (four months in total). Although the re-trained GBT (performed with 1.6 million transactions) achieved the best performance, it is 3000 times slower than the incremental approach.  Since re-training is too time-consuming, we aim to improve the accuracy and training time for learning incrementally. 
\section{Credit-Card Fraud Dataset Preparation}
We select a public, anonymized credit-card fraud dataset consisting of 284807 users' purchasing transactions that occurred during two days in September 2013 ~\cite{CCFD2016}. We may mention that credit-card fraud datasets are lacking due to data privacy. The 2-class dataset possesses 30 predictive features obtained with the feature extraction method PCA; the two features Time and Amount were not transformed as their actual values are essential. Time represents the seconds passed between the current and the first transactions in the dataset. After examining the dataset, only $0.172\%$ (492) of the transactions is fraudulent. We found duplicated fraud data (19) that we eliminate. We are in the presence of a highly imbalanced dataset where the count of legitimate data is much higher than the count of fraudulent data. In this situation, classifiers will be biased towards the normal class and will misclassify the fraud class. 

\smallskip

We divide the dataset into two chunks according to the transaction time: the first chunk is related to the first day and the second chunk to the second day. We use the first chunk to perform transfer learning, and the second chunk IFL. Next, using the stratified method, we split first chunk into 70\% training and 30\% testing and the second chunk into 70\% training, 15\% validation and 15\% testing. Table \ref{ccfd} (a) presents the original dataset's statistics. As given in Table \ref{ccfd} (a), the training chunks are highly imbalanced, therefore we adopt the famous over-samping method SMOTE with the new class distribution ratio of 1.3. We employ over-sampling to keep all the fraud data as they are rare events. For example, the training chunk2 has an imbalance ratio of 1:689 that we re-balance with a new ratio 1.3.  Table \ref{ccfd} (b) exposes the re-sampled train, validation and test chunks. Since Time is irrelevant for the classification task, we remove it.

\begin{figure}[H]
\centering
\includegraphics[width=1.00\linewidth]{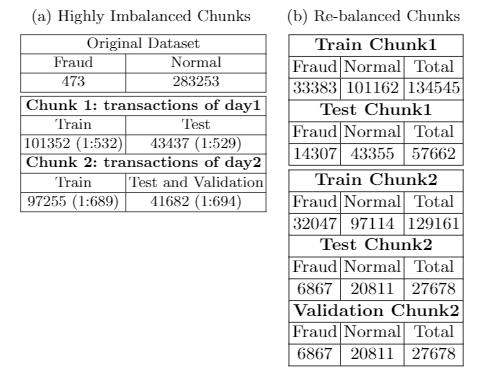}
\caption{Original and Re-Sampled Training and Testing Chunks}
\end{figure}  

\section{Transfer Learning and Incremental Feature Learning}
Transfer learning is usually adopted in image processing, where an existing classifier is used to detect new objects, as the classifier can efficiently extract new features from images.  In our fraud classification application, based on the initial model trained on the first-day transaction chunk, our algorithm utilizes the transformed features to train on the second-day chunk with the optimal number of hidden units.  Instead of re-training the entire model for each new chunk, we only train the newly added units and freeze all the previous layers. This strategy will significantly decrease the training time so that the classifier identifies fraud activities much faster, especially that our application requires real-time responses. The most apparent difference of incremental learning from conventional learning is that the former does not assume the availability of sufficient training data, as data are available over time ~\cite{DBLP:journals/npl/GuanL01}. So, based on these facts, improving learning for each newly supplied chunk is an excellent approach.\\

Algorithm 1 first conducts transfer learning and then IFL. We first build the initial network with four fully connected layers, where the first layer has the input features, the first hidden layer 500 units, the second hidden layer ten units, and the output layer one unit for the binary classification. The study ~\cite{DBLP:conf/smc/AnowarS20}, which validated an example-based incremental learning approach with the same credit-card dataset, determined that 500 units as the best number for the first hidden layer. So, we train this first network with the mentioned topology on the first chunk. Then, after removing the output unit, we refit the initial model using previous knowledge and second-day chunk.  Now,  the second hidden layer represents the transformed features (called tSubset), i.e., the new dimensional space more relevant to the target class because it is obtained with previous knowledge.  Here, tSubset contains the values corresponding to the transformed features. For example, assume we have 100,000 transactions in the second chunk. Using the refitted model, we convert each transaction of 29 values to ten new values. So, our transformed tSubset will contain 100,000 new data with 10 new features.  

\medskip
We actually extend the initial network with two sub-networks; the first NN considers the high-level features, and the second NN the low-level features. We leverage both of them to predict the output more efficiently.  Also, the study ~\cite{Wang2014EEGES} showed that creating sub-networks according to the features' relevancy can lead to a better outcome. We add the second sub-NN to improve more the fraud detection model performance. A similar idea is used in the incremental approach defined in ~\cite{DBLP:journals/npl/GuanL01} by connecting the original inputs directly to the output unit with the same motivation. Another paper ~\cite{DBLP:journals/isci/WangWYDC21} explained and compared the performance results of connecting the inputs to the output unit directly.  These are the reasons we use high-level features for creating the first sub-NN and then low-level features for creating the second sub-NN. Thus, we can gradually improve the accuracy using newly available data without forgetting the previous knowledge (stored in the previous weights). Moreover, the training time will be much lower as we only train the new chunk and one unit only per epoch. Our approach can be used for any number of chunks by appending a new hidden layer to the first sub-NN.

\medskip

Algorithm 2 shows how to extend a sub-NN with new hidden units and connections. The main challenge is determining the number of units that should be added to attain the best performance. In transfer learning, we add a predetermined number of layers and units to the previously trained model.  It can lead to fewer layers/units; therefore, the resulting model will return poor predictions. It can also lead to unnecessary layers/units, which only increases the learning time. Since the numbers of layers and units are predetermined, all the units are trained together and not one at a time like our proposed method. So, the training time will be very time-consuming. We determine the needed number of units in our work by adding and training them one by one. This approach is highly efficient computationally. Moreover, instead of computing the gradient descent for all the new units at once for each epoch, which is time-consuming, we utilize the ``patience" parameter as the early stopping callbacks. Based on the convergence threshold, this parameter checks whether the error is reducing or not in a certain number of epochs. In the experiments, if the accuracy does not increase in ten epochs, then the learning will stop. 

\newpage

\begin{footnotesize}
\begin{algorithm}[H]
{\bf Inputs:} trainChunk1, trainChunk2, validChunk2, threshold, nEpoch \\
\KwResult{optimal predictive model} 
\medskip
{\color{blue} (*Initial Model with chunk1*)}\\
iniNN $\gets$ build network with four fully connected layers;\\
iniModel $\gets$ train initNN on trainChunk1; \\

{\color{blue} (*Refitted Model with chunk2*)}\\
refModel $\gets$ delete output unit of initModel;\\
refModel $\gets$ feedforward trainChunk2 to refModel using past weights; \\

{\color{blue} (*Transformed Feature Sub-dataset*)}\\
tGroup $\gets$  extract set of transformed features;\\
tSubset $\gets$  create dataset w.r.t tGroup and trainChunk2;\\

{\color{blue} (*Incremental Feature Learning*)} \\
{\color{red} (*Sub-NN Training and Extension for Transformed Features*)} \\
subNN1 $\gets$ create network (2 hidden units and 1 output unit);\\
subNN1 $\gets$ connect subNN1 to output layer of refModel ;\\
model1  $\gets$ train subNN1 with tSubset by freezing past weights;\\
extModel1= {\bf Algorithm2}(model1, tSubset, validChunk2, threshold, nEpoch);\\

{\color{red} (*Sub-NN Training and Extension for Input Features*)} \\
subNN2 $\gets$ create network (2 hidden units);\\
subNN2 $\gets$ connect hidden units to output unit of extModel1 ;\\
model2 $\gets$ freeze weights of previous units except hidden and output units;\\
model2 $\gets$ train subNN2 with trainChunk2;\\
extModel2 $\gets$ {\bf Algorithm2}(model2, trainChunk2, validChunk2, threshold, nEpoch);\\
return extModel2;\\
\caption{Transfer Learning and Incremental Feature Learning}
\end{algorithm}

\begin{algorithm}[H]
{\bf Inputs:} {model; trainData, validData, threshold, nEpoch}\\
\KwResult{extended model} 
\medskip
preValAcc $\gets$ $0$;
\While{True}{
curValAcc $\gets$ compute accuracy of model using validData;\\
\While{(curValAcc is converging with patience of nEpoch)}{
     model $\gets$ train non-frozen weights with trainData;\\
     curValAcc $\gets$ compute accuracy of model using validData;\\
}

\If{(curValAcc - preValAcc $<$ threshold) and (preValAcc $\neq$ 0)}{
 Break;
}
preValAcc $\gets$ curValAcc;
model $\gets$ add new hidden unit to hidden layer;\\
model $\gets$ freeze weights of previous units except new hidden and output units;}
return model;
\caption{Model Extension with New Hidden Units until No Convergence}
\end{algorithm}
\end{footnotesize}

\subsection{A Concrete Example}
Based on the credit-card fraud dataset, we illustrate the behavior of our new algorithm in Figures 2, 3, and 4, where the green color denotes new connections and units, and red the frozen part. Figure 2 creates the initial network, trains it on chunk1, and after deleting the output unit, utilizes previously learned weights, and creates the refitted model by training with chunk2. Figure 3 adds the first sub-NN and keeps training on data of the ten features by including new hidden units and connections and freezing past parameters until the performance is not converging anymore.  Figure 4 creates the second sub-NN using the original features, connects it to the past output unit, freezes previous weights, and trains it on chunk2 until the model does not convergence anymore.

\begin{figure}[h]

\centering
\subfloat[create initial model on chunk1]{
\includegraphics[width=0.40\linewidth]{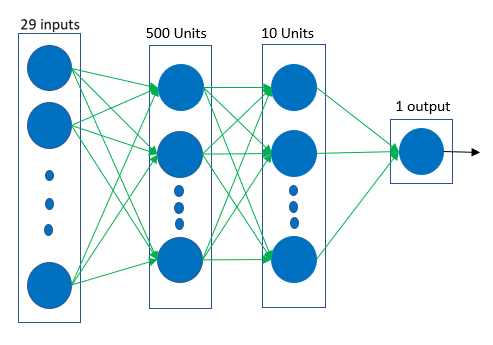}
}
\subfloat[transform inputs on chunk2 using past weights]{
\includegraphics[width=0.36\linewidth]{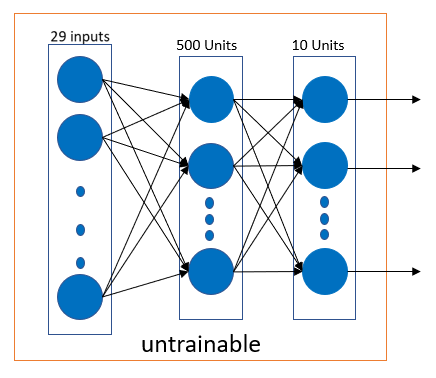}
} 
\caption{Transfer Learning}

\end{figure}

\begin{figure*}[h]
\centering
\subfloat[train subNN1 \& freeze past NN]{
\includegraphics[width=0.45\linewidth]{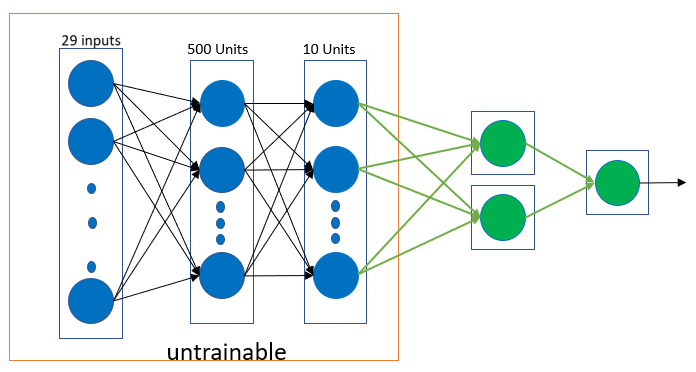}
}
\subfloat[add hidden unit(s) until no convergence]{
\includegraphics[width=0.50\linewidth]{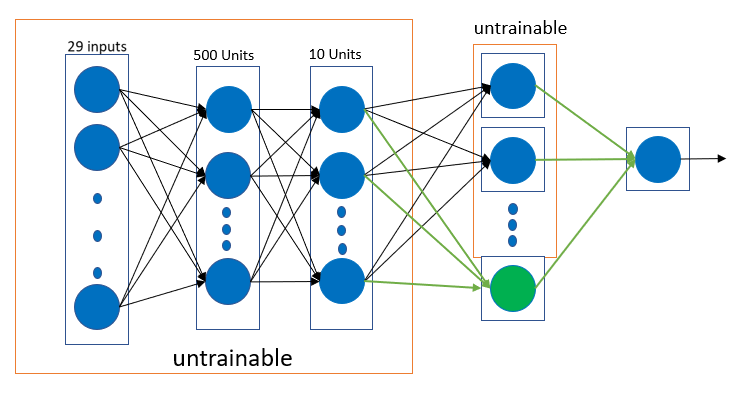}
} 
\caption{Network training and expansion using transformed features}

\end{figure*}

\begin{figure*}[h]
\centering
\subfloat[connect subNN2 \& freeze past NN]{
\includegraphics[width=0.45\linewidth]{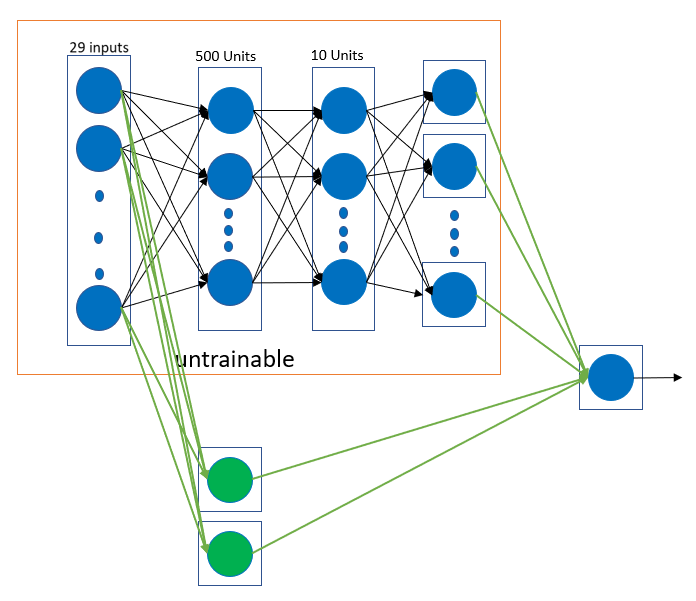}} 
\subfloat[add hidden unit(s) until no convergence]{
\includegraphics[width=0.45\linewidth]{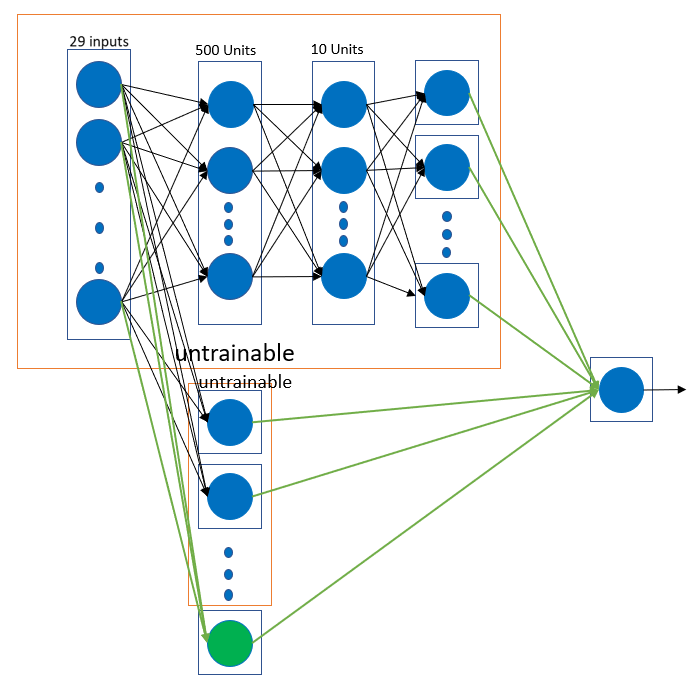}}
\caption{Network training and expansion using original inputs}

\end{figure*}

\section{Validation}
\subsection{Experiment Setup}
Since the 29 input features of our fraud dataset have different ranges of values and with significant discrepancies between the features, we first normalize them into the same range. Then, we train the initial network (four fully connected layers) on chunk1 data using different ranges to determine the most appropriate feature scale. The range of [-5, +5] returns the best performance. Furthermore, we tune the network hyperparameters, such as the learning rate to 0.001, batch size to 1024, epoch to 100, convergence threshold to 0.01, and patience to 10. We utilize five quality metrics for evaluating the predictive performance: Precision, Recall, F1-score, FNR, and Time (in seconds). Due to the randomness of NNs, we run ten training sessions and ten corresponding testing sessions. We consider the average of test sessions for comparing different fraud detection models.  

\subsection{Feature Transformation}
Using the trained model's knowledge obtained with the first-day transaction chunk, we transform the input features by feed-forwarding the second-day chunk to the first model. The new features are now more relevant to the target class, and after checking their correlation values, we find them much higher. Figure \ref{disNew} exposes the distribution of the ten transformed features.

\begin{figure}[h]
\centering
\includegraphics[width=0.9\linewidth]{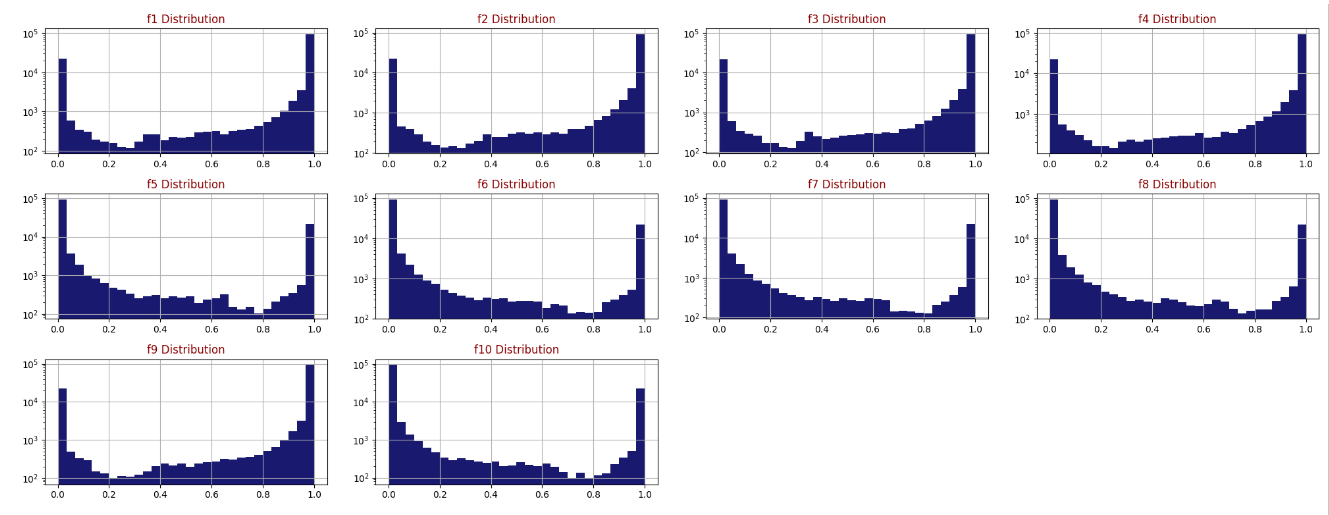}
\caption{Distribution of Transformed Features}

\end{figure}

\subsection{Performance Evaluation}
We first build two initial models, the first one is trained on train chunk1 and assessed on test chunk1 (presented in Table 1), and the second one using train chunk2 and test chunk 2 (presented in Table 2). According to the 10-round testing results of the two models, F1-score decreased by 4.6\% for day2 transactions. This expected decrease is due to the much higher number of instances in chunk1 (60\% of data) compared to chunk2 (40\% of data). Moreover, by preserving past knowledge, we re-fit the initial model (trained on chunk1) using train chunk2 to conduct transfer learning and evaluate its accuracy on test chunk2.  In this case, F-score decreased by 5.1\%. Changes in data patterns may have caused this decrease \cite{DBLP:conf/dsaa/LebichotPBSHO20}. For instance, if a concept drift occurred in the credit card data, the model performance decreases ~\cite{DBLP:journals/jnca/AbdallahMZ16}.  Lastly, we develop the final optimal model using our incremental feature learning approach. F1-score improved by 9\% using only one chunk.  Also, Recall, a significant metric in fraud detection, is augmented with 9.8\%, which means we have fewer false negatives in the second chunk. Also, as we are training one hidden unit at a time, the training time is less than the re-fitting time. Therefore, we can conclude that the final model outperforms the initial and re-fitted models using only one chunk. 

Another essential metric for fraud detection is the False Negative Rate (FNR), which refers to the rate of fraudulent transactions detected as normal ones. According to the average of the ten experiments, as exposed in Figure 6, the initial-model FNR on chunk1 is $0.149$, the initial-model FNR on chunk2 is $0.219$, the re-fitting-model FNR is $0.228$, and the final-model FNR is $0.13$. These values show that our model caches more fraudulent cases when fed with new chunks, reducing FNR over time. 

Financial transactions cannot be stored for a long time by the fraud classifiers because of privacy issues.  In this case, we train the model incrementally chunk by chunk and discard each chunk right away. As we observe, the proposed approach can enhance the performance using each chunk at a time and with less computational cost.

\begin{figure}[H]
\centering
\includegraphics[width=1.00\linewidth]{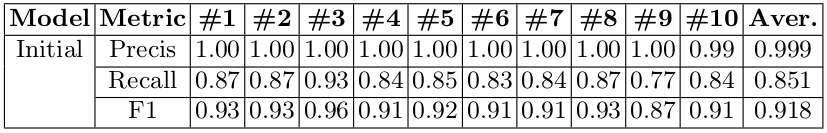}
\caption{Training and Testing on Day1 Transactions}

\end{figure}  

\begin{figure}[H]
\centering
\includegraphics[width=1.00\linewidth]{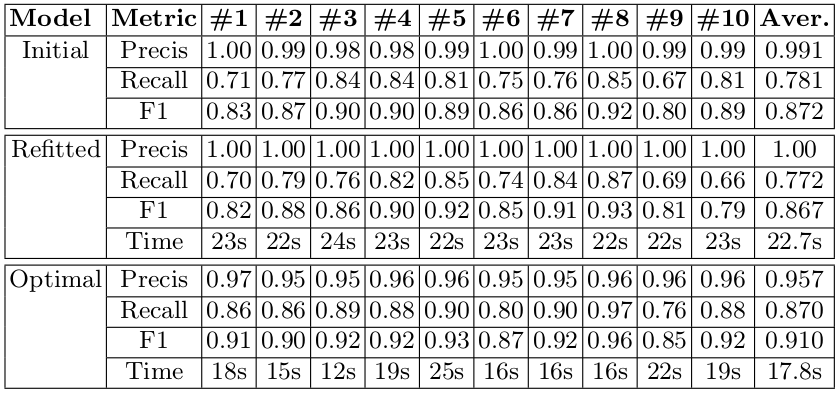}
\caption{Training and Testing of Different Models on Day2 Transactions}

\end{figure}

\begin{figure}[H]
\centering
\includegraphics[width=0.65\linewidth]{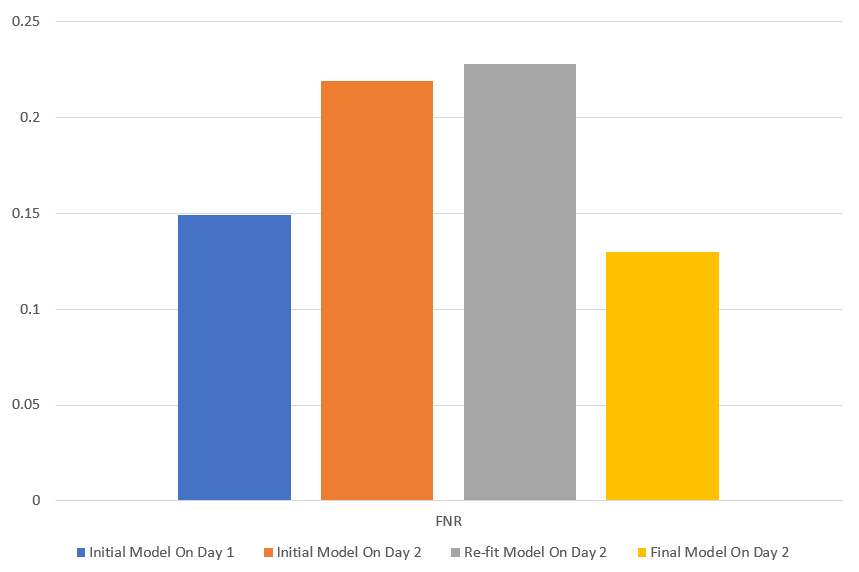}
\caption{FNR Comparison on the Four Incremental Models }

\end{figure}  

\section{Conclusion}
For tackling the actual behavior of the credit-card fraud detection environment, we introduced a new classification algorithm that adjusts gradually and efficiently to incoming chunks based on transfer learning and IFL. Transfer learning preserves past knowledge and utilizes it to adapt to subsequent chunks.  IFL extends the network topology incrementally by finding the optimal number of hidden units for detecting fraud in each chunk. Our hybrid approach improves the predictive accuracy without the necessity of accumulating a substantial volume of data and spending too much time on training. Our new approach can be employed on everyday credit card transactions to prevent the performance from decreasing using current and past knowledge.

\bibliography{Paper}
\bibliographystyle{alpha}
\end{document}